\documentclass[letterpaper, 10 pt, conference]{ieeeconf}
\IEEEoverridecommandlockouts     
\usepackage{bbold}
\usepackage{amsmath}
\DeclareMathOperator*{\argmax}{argmax}

\usepackage{graphicx}
\usepackage{tabularx}
\newcolumntype{Y}{>{\centering\arraybackslash}X}

\usepackage[ruled]{algorithm2e}
\SetKw{Continue}{continue}
\SetArgSty{textnormal}

\makeatletter
\let\NAT@parse\undefined
\makeatother
\usepackage[numbers]{natbib}

\title{\LARGE \bf On Enhancing Ground Surface Detection from Sparse Lidar Point Cloud}

\author{Bo Li$^{1}$
\thanks{*This work was done at TrunkTech. Contact: \texttt{prclibo@gmail.com}}}

\date{}

\begin{document}

\maketitle

\begin{abstract}
Ground surface detection in point cloud is widely used as a key module in autonomous driving systems. Different from previous approaches which are mostly developed for lidars with high beam resolution, e.g. Velodyne HDL-64, this paper proposes ground detection techniques applicable to much sparser point cloud captured by lidars with low beam resolution, e.g. Velodyne VLP-16. The approach is based on the RANSAC scheme of plane fitting. Inlier verification for plane hypotheses is enhanced by exploiting the point-wise tangent, which is a local feature available to compute regardless of the density of lidar beams. Ground surface which is not perfectly planar is fitted by multiple (specifically 4 in our implementation) disjoint plane regions. By assuming these plane regions to be rectanglar and exploiting the integral image technique, our approach approximately finds the optimal region partition and plane hypotheses under the RANSAC scheme with real-time computational complexity.

\end{abstract}

\section{Introduction}

Recent development of lidar products brings significant progress for autonomous driving. Velodyne releases dense beam product VLS-128 and the price of VLP-16 is largely reduced. New manufacturers like Hesai and RoboSense keep releasing new products with 16, 32 or 40 beams to the market in the past two years. Based on the application scenario, autonomous driving systems can accordingly select suitable lidars for the perception module. For example, HDL-64 is replaced by 32- or 40- beam lidars for cost reduction in many recent passenger vehicle systems. In low-speed applications like small delivery vehicles, VLP-16 can be used for further cost compression.
Along with the development of lidar hardware, research on environment perception from lidar point cloud data draws increasing attention. Besides the traditional geometry based object detection approaches, deep learning based object detection techniques are also transplanted to point cloud data and have achieved promising performance \cite{Li2016, yan2018second}. In addition to object detection, ground surface detection from point cloud is also an important topic in autonomous driving. The value of ground surface detection comes from two aspects: 1) ground detection can be interpreted as the dual problem of generic obstacle detection; 2) segmented ground surface can be directly used as the drivable area for vehicle motion planning.

A variety of previous works have been drawn on ground surface detection but mostly designed to work within regions with high density of points (usually within $30$m from a HDL-64). Such approaches are insufficient for many autonomous driving applications. Passenger vehicles running at $70$km/h should detect obstacle at least $40\sim50$m away and low speed delivery vehicles should do up to $30$m. Note that the point density of VLP-16 at $30$m is simliar to that at $40\sim50$m for HDL-64. Many previous approaches do not apply to such density. To solve this problem, we proposes an approach to detect ground surface from regions with sparse points.

\begin{figure}
    \centering
    \begin{tabular}{cc}
        \includegraphics[width=0.2\textwidth]{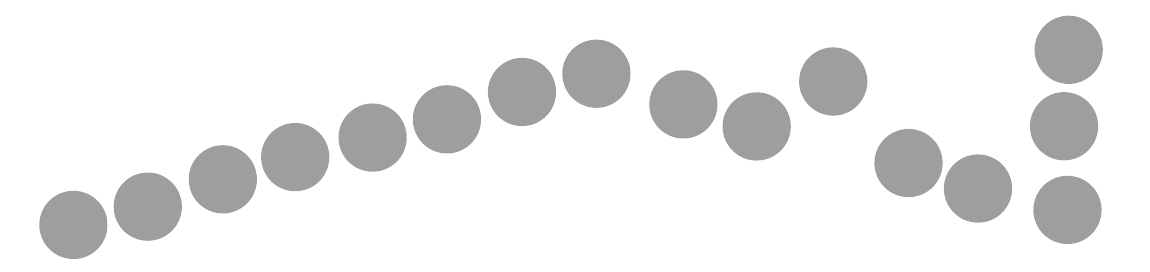} & \includegraphics[width=0.2\textwidth]{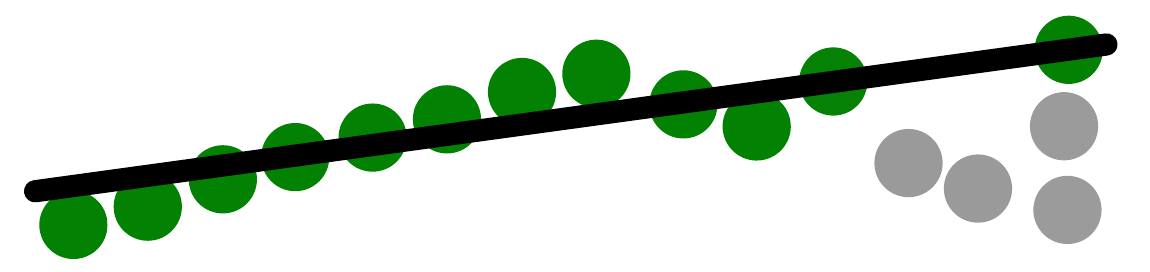} \\
        (a) & (b) \\
        \includegraphics[width=0.2\textwidth]{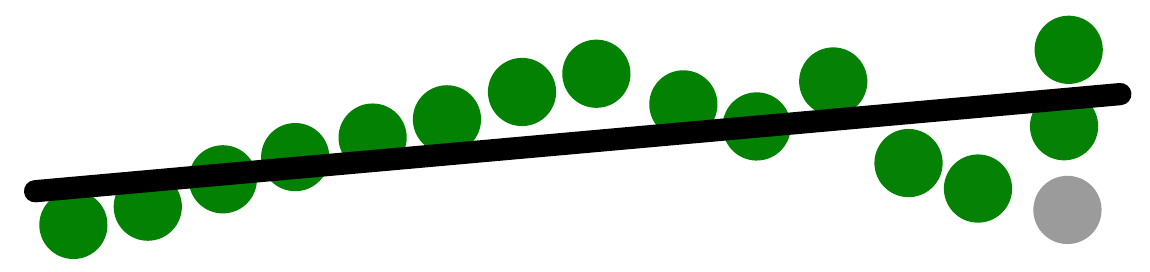} & \includegraphics[width=0.2\textwidth]{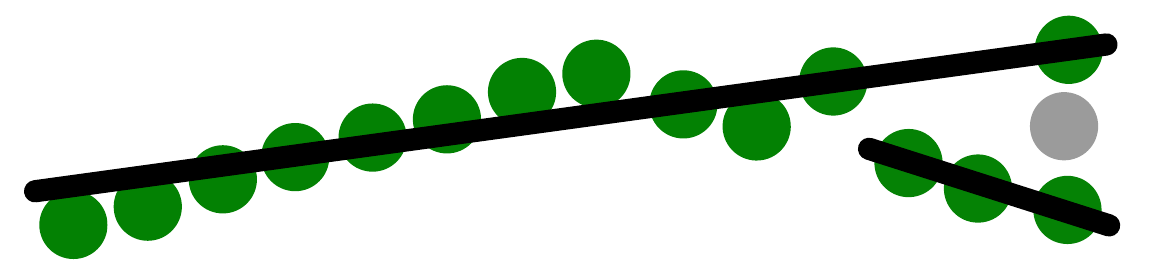} \\
        (c) & (d)\\
        \includegraphics[width=0.2\textwidth]{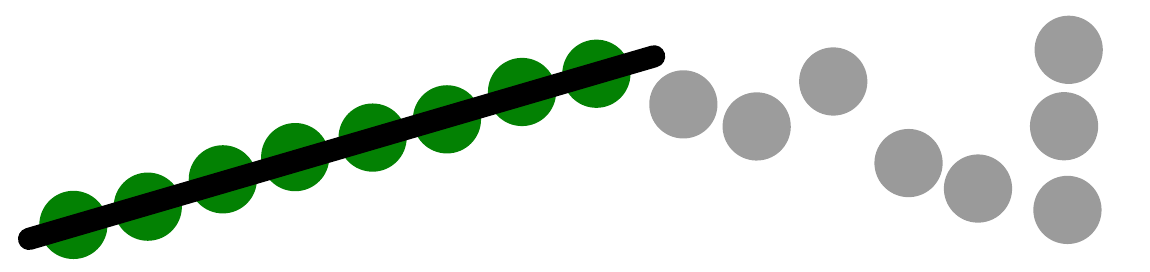} & \includegraphics[width=0.2\textwidth]{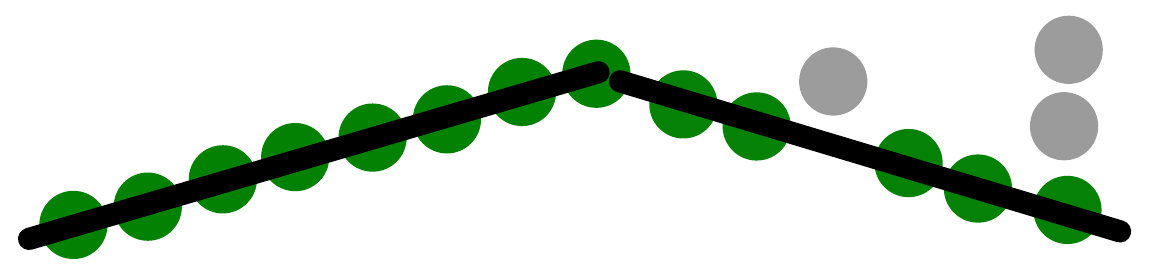} \\
        (e) & (f)
    \end{tabular}
    \caption{Toy examples of some plane fitting approaches illustrated in a 2D side view. Black lines denote planes. Grey data points are colored green if taken as inliers. (a) Raw data. (b) Plane fitted by vanilla RANSAC. (c) Same as (b) but with a larger inlier threshold. (d) Two planes greedily fitted by vanilla RANSAC. (e) Plane fitted by RANSAC with tangent based inlier verification. (f) Two disjoint plane fitted by RANSAC with tangent based inlier verification and optimal partition search.}
    \label{fig:toy-sample}
\end{figure}

\section{Related Works}
\label{sec:related}

We briefly revisit the existing approaches on ground surface detection from point cloud by roughly categorizing these approaches as model-fitting and labeling methods.

\paragraph{Model-fitting} A typical series of previous approaches fit ground surface by planes. Vanilla RANSAC based plane fitting can be enhanced by taking point-wise normal into account for hypothesis verification \cite{Schnabel2007}. \citet{Chen2017} convert plane fitting to line fitting in a dual space for better performance. By assuming known lidar height and longitudinal road slope, the ground plane representation can be reduced to one parameter \cite{Choi2014}. However, in realistic applications, ground surface is usually not perfectly planar, in which case a single model cannot fit well. \citet{Zermas2017} uniformly partition the ground surface to three parts and fit planes respectively. \citet{Himmelsbach2010} divides ground surface into angular grid and fits lines instead of planes for each angular direction respectively. These improvement increases the fitting flexibility, but some partition segments might contain too few points to fit when the point cloud is sparse.

\paragraph{Labeling} The labeling methods exploit local features and do not rely on global geometry model. \citet{Petrovskaya2008, Moosmann2009, Bogoslavskyi2016} search disconnection between points and label ground points by region growing. \citet{Nitsch2018} uses point-wise normal and normal quality as classification features. If assuming lidar mounted perfectly vertical, adjacent ring distance can also be used as classification feature \cite{Choi2013}. These features can be extracted from dense point cloud regions, for example HDL-64 within $30$m or VLP-16 within $10$m. However, it is difficult to extract such features from sparse point cloud region which is studied in this paper. Inspired by the recent development of deep learning on point cloud data, Convolutional Neural Network (CNN) is also used for feature extraction and ground segmentation \cite{velas2018cnn, lyu2018real}. Besides feature design, higher order inference approaches are also studied to guarantee the ground surface smoothness. Representative techniques include Markov Random Field (MRF) \cite{Zhang2015, Guo2011, Byun2015}, Conditional Random Field (CRF) \cite{rummelhard2017ground} and Gaussian Process (GP) \cite{Chen2014a, Douillard2011}. \citet{rummelhard2017ground, Douillard2011} use height as classification feature in unary item. \citet{Zhang2015} relies on significant local height change to distinguish obstacles to define unary cost for obstacles. \citet{Byun2015, Guo2011, Chen2014a} build features based on local surface gradients. \citet{Douillard2011} combines plane fitting and GP to overcome ground point noise. Besides the difficulty of normal/gradient estimation in sparse point cloud, one common problem for higher order inference approaches is the ambiguity between sloped ground and far obstacles in very sparse point cloud.









\section{Approaches}
\subsection{Tangent Based Inlier Verification}

\begin{figure}
    \vspace{0.5em}
    \centering
    \includegraphics[width=0.48\textwidth]{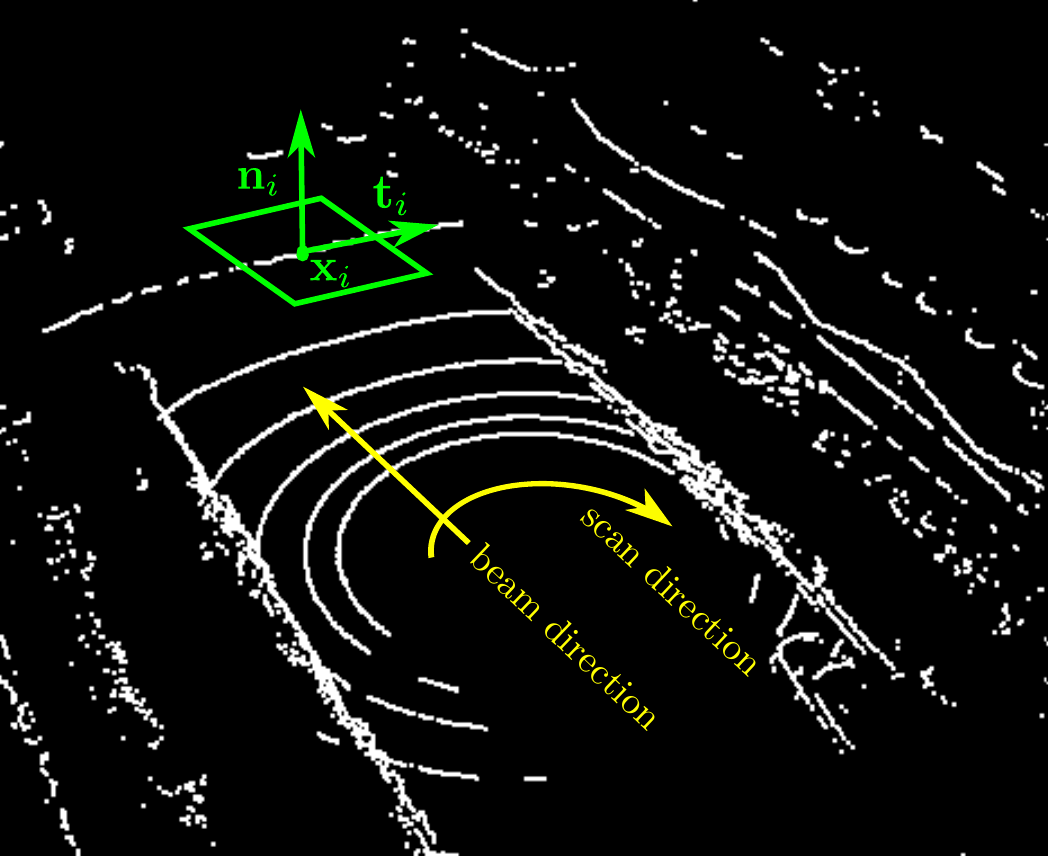}
    \caption{A sample point cloud captured by a Velodyne VLP-16 lidar. Yellow axes denote the scan direction corresponding to the column direction in the range image and the beam direction corresponding to the row direction.}
    \label{fig:coordinates}
\end{figure}

In a vanilla RANSAC based plane detection approach, inliers are verified according to point-plane distance by
\begin{equation}
    | \mathbf{n}_\text{pl}^\top \mathbf{x}_i + d | < \epsilon
\end{equation}
where $(\mathbf{n}_\text{pl}^\top, d)$ is the plane coeffients and $\mathbf{x}_i$ the point coordinates. $\epsilon$ is the inlier threshold. Such criterion is not sufficient to distinguish false-positive in many cases. Figure \ref{fig:toy-sample}b shows an example where a fitted plane is deviated by false-positive inliers. The false-positive inliers belong to the right vertical wall plane but have close distance to fitted ground plane.

To avoid such false-positive inliers, verification can be enhanced if point-wise normal is available \cite{Schnabel2007}. This criterion can be denoted as:
\begin{equation}
\begin{aligned}
    | \mathbf{n}_\text{pl}^\top \mathbf{x}_i + d | &< \epsilon \\
    \text{arccos} (\mathbf{n}_\text{pl}^\top \mathbf{n}_i) &< \delta
\end{aligned}
\label{eq:criterion2}
\end{equation}
where $\mathbf{n}_i$ denotes the normal estimated at point $\mathbf{x}_i$ and $\delta$ the angle threshold for normal difference. A variety of previous approaches have been proposed to estimate point normal in point cloud, e.g. \cite{badino2011fast}. 

As mentioned in Section \ref{sec:related}, point-wise normal estimation is not always reliable for sparse point cloud captured by lidar in autonomous driving. In this paper we take point cloud captured Velodyne VLP-16 as a representative example as is shown in Figure \ref{fig:coordinates}. Points in the green plane patch belong to ground plane but are difficult to estimate normals. The neighbor points in the green patch are approximately distributed on a straight line. Thus it is ill-posed to estimate a plane and its normal locally from these points.

In this paper we enhance the inlier verification in sparse point cloud without estimating point-wise normal. Consider the well-known 2D range scan organization of lidar point cloud used in many previous research \cite{Bogoslavskyi2016, badino2011fast}. We observe that regardless of the sparsity in the vertical beam direction, point density along the horizontal scan direction is always high. For a Velodyne HDL-64 or VLP-16 spinning at 10Hz, the azimuth angle difference between adjacent points from a same beam is less than $0.2^{\circ}$. If interpret points produced by the same beam as a curve, such density is sufficient to estimate the tangent of the curve with good precision. Furthermore, since this curve is embedded on an object (ground) surface locally, the estimated tangent vector is also one of the tangents of the surface plane. Note that for high beam resolution lidar it is also possible to estimate tangent along the vertical beam direction and thus the point normal is determined. Previous works like \cite{badino2011fast, Guo2011} make use of this property to accelerate normal/gradient estimation in range scans. With only one valid tangent on a plane in our cases, we can relax the RANSAC criterion (\ref{eq:criterion2}) based on the perpendicularity between the tangent and the plane normal to estimate:
\begin{equation}
\begin{aligned}
    | \mathbf{n}_\text{pl}^\top \mathbf{x}_i + d | &< \epsilon \\
    |\pi / 2 - \text{arccos} (\mathbf{n}_\text{pl}^\top \mathbf{t}_i) | &< \delta
\end{aligned}
\label{eq:criterion3}
\end{equation}
where $\mathbf{t}_i$ is the estimated curve tangent at point $\mathbf{x}_i$. $\mathbf{t}_i$ can be estimated by taking numerical differentiation of points from a same beam.

Figure \ref{fig:toy-sample}e provides a toy example of RANSAC with tangent based outlier rejection in a 2D view. Points on the left ground have different tangent direction with those on the right ground or the vertical wall, which avoids the fitted plane to compromise over different surfaces.

\subsection{Disjoint Multiple Plane Fitting}
\label{subsec:disjoint}
Leveraging local point-wise tangents for inlier verification enhances the robustness of RANSAC plane fitting. However, as mentioned in Section \ref{sec:related}, fitting one plane is still not sufficient to precisely model the ground surface of real road which is not perfectly planar. Take Figure \ref{fig:toy-sample} as an example again. Fitting a dominant ground plane with small inlier threshold leaves remaining ground points as false-negatives, which is the case shown in Figure \ref{fig:toy-sample}b and \ref{fig:toy-sample}e. On the other hand, increasing inlier threshold will mistaken outliers as false-postive, which is the case shown in Figure \ref{fig:toy-sample}c. For each time of laser emission in VLP-16, only $6 \sim 7$ out of $16$ points will hit the ground. Thus partitioning point cloud into fan-like thin segments as proposed in \cite{Himmelsbach2010, Chen2014a} results in too few points for further modeling. In addition, it is difficult for MRF or GP to distinguish whether the elevation difference of a far neighbor points is due to obstacle or gradual ground elevation increase.

In order to model non-planar road surface without introducing too much flexibility or hyper-parameters, we propose to use multiple horizontally disjoint plane segments to fit the ground surface points in this paper. The property of disjointness means that the ground surface at every horizontal position has only one layer. Figure \ref{fig:toy-sample}d provides a simple example of naively fitting multiple planes in a greedy strategy. Points of two different planes at the right of the scenes are taken as inliers respectively, which causes the ground surface to have two ambiguous layers. \citet{Zermas2017} partitions point cloud by fixed distance at the longitudinal direction and fit planes independently for disjointness. However, it is impossible to find a fixed partition which is suitable for different surface deformation.

Instead of fixing partition, we propose to search for the best partition to fit multiple disjoint plane for each point cloud respectively. We first illustrate a general formulation of this problem and its computational complexity. Suppose we have $P$ partition cases over the horizontal plane and the point cloud contains $N$ points. The $p$-th partition divides the horizontal plane into $S_p$ segments. Based on the RANSAC plane fitting strategy, we first generate $M$ plane hypotheses. The general formulation of our problem searches for best partition and best plane for each segment which maximizes the total inlier number:
\begin{equation}
\begin{aligned}
    \max_{0 \leq p < P} \max_{m_0, \cdots, m_{S_p - 1}} \sum_{s = 0}^{S_p - 1} \sum_{i = 0}^{N - 1} f(i; m_s) \land g_p(i; s)\\
    = \max_{0 \leq p < P} \sum_{s = 0}^{S_p - 1} \max_{m_s}  \sum_{i = 0}^{N - 1} f(i; m_s) \land g_p(i; s)\\
\end{aligned}
\label{eq:general-objective}
\end{equation}
In this above formulation, $s$ is used to index a segment region on the horizontal plane given a partition indexed by $p$ and $m_s \in \{0, \cdots, M - 1\}$ is used to index the plane hypothesis assigned to segment $s$. $f(i; m_s)$ denotes whether the $i$-th point is a inlier of plane $m_s$ and $g_p(i; s)$ denotes whether this point is within segment region $s$ in a partition $p$:
\begin{equation}
\begin{aligned}
    f(i; m) &= 
    \begin{cases}
        1 & \text{point $i$ is an inliers for plane $m$}\\
        0 & \text{otherwise}
    \end{cases}\\
    g_p(i; s) &=
    \begin{cases}
        1 & \text{point $i$ in segment region $s$ in partition $p$}\\
        0 & \text{otherwise}
    \end{cases}
\end{aligned}
\label{eq:f-and-g}
\end{equation}
(\ref{eq:f-and-g}) can be computed in a constant computational complexity time. Denote $S$ as a superior of $S_p$. The overall computational complexity of exhausted search over objective (\ref{eq:general-objective}) is $O(P S M N)$. Such complexity is not tractable for real-world application. Specifically, even if we limit the segment number and discretize partition cases, the number of general partition cases on a given plane region is still not tractable.

We reduce the complexity by first assuming that partitions over the horizontal plane are always aligned with the $xy$ coordinate axes s.t. each partition segment is a rectangle area aligned with $xy$ axes. This assumption helps simplify the inner part of (\ref{eq:general-objective}), denoted as $h_p(m, s) = \sum_{i = 0}^{N - 1} f(i; m_s) \land g_p(i; s)$ for short.

Bound the whole point cloud within a horizontal square, which is uniformly divided into $B \times B$ bins. Given a plane $m$, denote $b_m(c, r)$ with $0 \leq c, r < B$ as the number of inliers of plane $m$ which fall in the bin $(c, r)$. Denote $I_m(c, r)$ as the integral image of $b_m(c, r)$. If we round a segment $s$ as its closest rectangle aligned with the bin grid, $h_p(m, s)$ can be easily approximated from $I_m$. Denote the rectangle segment $s$ by its top left and bottom right corners $(c_0, r_0)$, $(c_1, r_1)$. $h_p(m, s)$ is obtained by:
\begin{equation}
    h_p(m, s) \approx I_m(c_1, r_1) + I_m(c_0, r_0) - I_m(c_0, r_1) - I_m(c_1, r_0)
\end{equation}
With this approximation, we reduce the complexity to $O(P S M)$ given precomputed $I_m$. The complexity of its precomputation is $O( (N + B^2) M)$ (see the first for loop in Algorithm \ref{alg:alg}).

\begin{figure}
    \vspace{0.8em}
    \centering
    \includegraphics[width=0.4\textwidth]{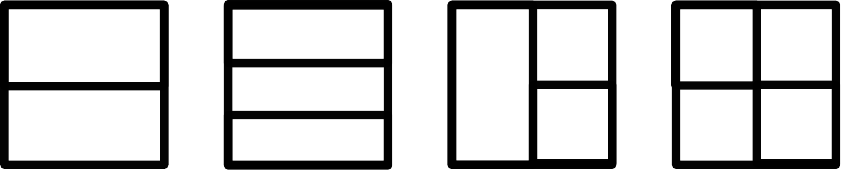}
    \caption{Samples of partition series. In this paper we restrict that partitions are always made up of rectangles. We name the 4-th partition series the cross-shape-partition. See Section \ref{subsec:disjoint}}.
    \label{fig:partition}
\end{figure}

Next we consider the partition formulation. Figure \ref{fig:partition} shows some simplest partition series aligned with $xy$ axes. Note that \cite{Zermas2017} can be interpreted to use the second series of partition (three parallel rectangles) in Figure \ref{fig:partition} and fix partition position s.t. $P = 1$ and $S = 3$ for complexity reduction. In the proposed approach, we use the forth cross-shape-partition series shown in Figure \ref{fig:partition}. This partition series compromises well over different surface deformation without introducing too much flexibility. Thus $S$ is fixed as $4$. Since we approximate partition segment to align with $B \times B$ bins, the cross-shape-partition series contains $B^2$ partition cases, i.e. $P = B^2$. These $B^2$ cases are equivalent to enumerating the cutting cross over the $B^2$ bin centers. Therefore, the complexity $O(P S M)$ is further reduced to $O(B^2 M)$, omitting contant $S$ (see the second and third for loop in Algorithm \ref{alg:alg}). Combining with the precomputation procedure, the overall complexity is bounded by $O((N + B^2) M)$. A Velodyne VLP-16 spinning at 10Hz generally produces approximately $30$k points per frame. For a $80$m $\times$ $80$m point cloud scene, bin size of $1$m results in $B^2 = 6.4$k bins and provides good partition precision. Algorithm \ref{alg:alg} illustrates the pesudo-code for the overall procedure. Note that to avoid overfitting on small outlier planes, we restrict each segments to have at least $T$ inliers. 

\begin{algorithm}
    Sample $M$ RANSAC plane hypotheses\\
    
    $b[:, :] \gets 0$\\
    \For {$m\gets0$ \KwTo $M - 1$}{
        \For {$i\gets0$ \KwTo $N - 1$}{
            $(c, r) \gets$ belonging bin index for $\mathbf{x}_i$\\
            $b[c, r] \gets b[c, r] + f(i; m)$
        }
        $I_m \gets$ integral image of $b_m$\\
    }
    $S \gets 4$\\
    $A[:, :, :, :] \gets 0$\\
    \For {$0 \leq c < B, 0 \leq r < B$}{ 
        \For {$m \gets0$ \KwTo $M - 1$}{
            \For {$s \gets0$ \KwTo $S - 1$}{
                $p \gets r * B + c$\\
                $A[c, r, s, m] \gets h_p(m, s)$\\
            }
        }
    }
    \For {$0 \leq c < B, 0 \leq r < B$}{
        $m[:] \gets 0$\\
        \For {$s \gets0$ \KwTo $S - 1$}{
            $m[s] \gets \argmax_m A[c, r, s, m]$
        }
        curr\_sum $\gets \sum_p A(c, r, s, m[s])$\\
        \If {\texttt{curr\_sum} $>$ \texttt{best\_sum} $\land  \min m[s] > T$}{
            \texttt{best\_sum} $\gets$ \texttt{curr\_sum}\\
            update best $c, r, m[:]$
        }
    }
    \caption{RANSAC based ground surface fitting assuming a cross-shape-partition of horizontal plane}
    \label{alg:alg}
\end{algorithm}

\section{Experiments}
In our experiments we compare the proposed approach with three representative previous works.
\begin{itemize}
    \item Vanilla RANSAC fits a single plane over the whole point cloud.
    \item MRF implements the Markov Random Field (MRF) approach proposed in \cite{Zhang2015}, where a MRF is constructed over a cylindrical grid map to estimate the ground height.
    \item LPR implements the ground surface fitting module proposed in \cite{Zermas2017}, where the Lowest Point Representatives (LPR) are used to initialize iterative ground plane fitting for a fixed partition.
\end{itemize}

\subsection{Implementation Details}
In the experiments, the above four approaches are implemented in C++ without parallel optimization.
\begin{itemize}
    \item Point cloud captured by Velodyne VLP-16 within a radius of $40$m is used as input. The extrinsics between lidar and vehicle frame are known to counteract the lidar mounting error. 
    \item The point cloud is sampled over a horizontal grid with resolution $0.1$m, producing approximately $8$k points.
    \item Vanilla RANSAC and our proposed method both generates $M=200$ plane hypotheses.
    \item Inlier threshold of point-plane distance for Vanilla RANSAC, LPR and our proposed method is $0.2$m. For MRF, points which are $0.2$m higher than the estimated ground height are classified as obstacles.
    \item The proposed method uses a grid with size of $B = 80$, i.e. bin size of $1$m.
    \item MRF uses same cylindrical grid map resolution, cost function parameters and number of iterations with the \cite{Zhang2015}. The unrevealed binary cost weight and binary cost superior is selected empirically.
    \item LPR uses same algorithm parameters including the number of plane fitting iterations and the candidate number of LPR with \cite{Zermas2017}. The scene uniformly divided to three parts along the longitudinal direction.
\end{itemize}

\subsection{Results Analysis}
\label{subsec:results}

\begin{figure*}
    \vspace{0.8em}
    \centering
    \includegraphics[width=\linewidth]{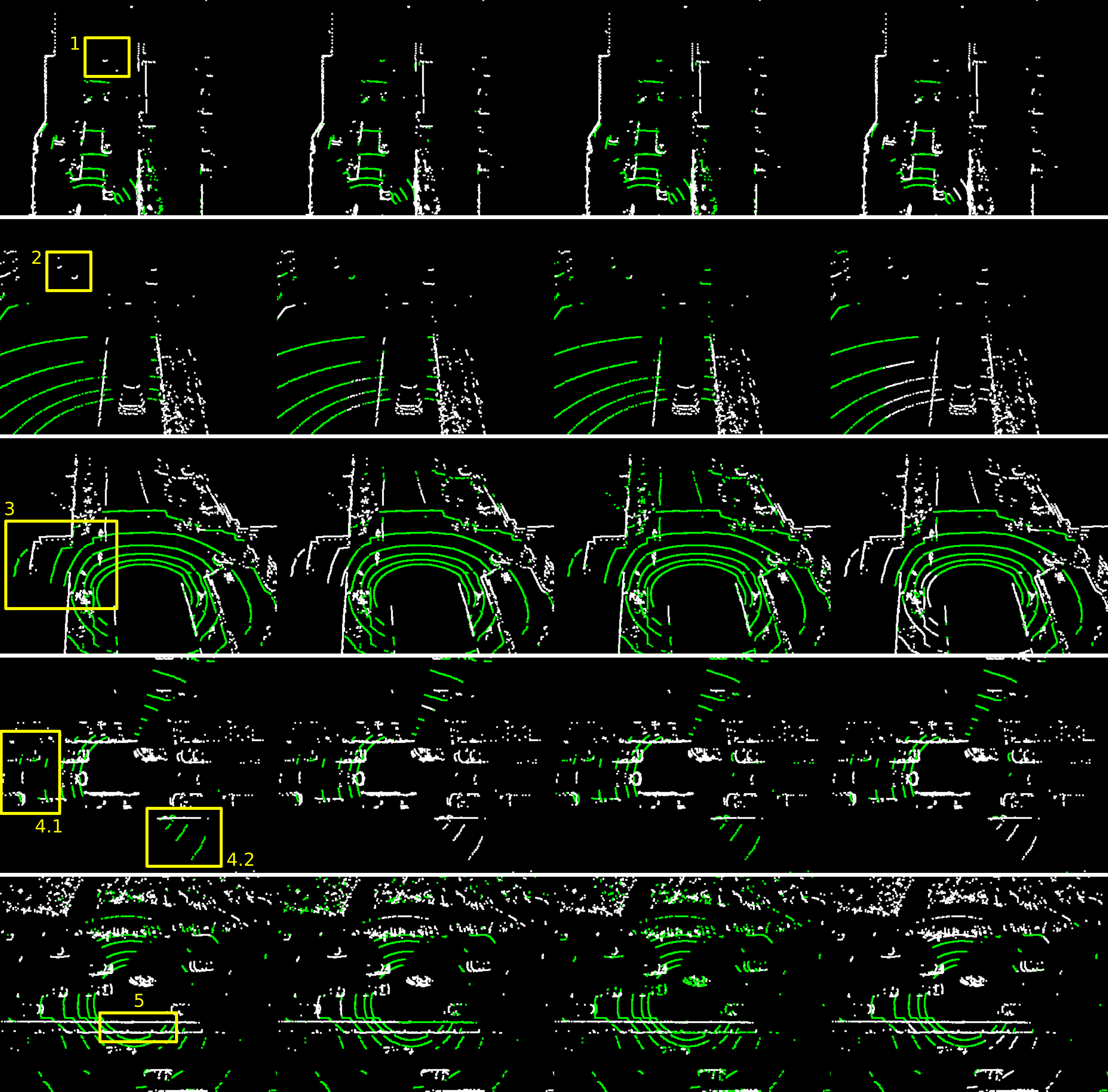}
    \begin{tabularx}{\linewidth}{YYYY}
         Proposed & Vanilla RANSAC & MRF \cite{Zhang2015} & LPR \cite{Zermas2017} \\
    \end{tabularx}
    \caption{Sample results produced by approaches in our experiements. Detected ground points are labeled as green. See Section \ref{subsec:results} for analysis.}
    \label{fig:comparison}
\end{figure*}

Figure \ref{fig:comparison} shows some representative results of the compared approaches. The proposed approach shows promising performance enhancement on sparse point cloud. We discuss the following cases marked by yellow boxes which are considered difficult for previous approaches.

\paragraph{Far obstacles} Box 1 and Box 2 mark typical examples of far obstacles approximately $35$m away from the VLP-16. In such cases, obstacles (e.g. a short scan segment on vehicle) are higher than the ground plane but have no nearby neighbor ground points. Thus MRF can not distinguish if far points are from obstacles or a part of sloped ground. Vanilla RANSAC also mistakens far obstacles as ground points in Box 1 and Box 2. This failure is similar to the case of Figure \ref{fig:toy-sample}c.

\paragraph{Sloped ground region} Box 3 shows an example of a sloped lane (with a gate). Vanilla RANSAC of a single plane fails to fit the sloped region as was expected. Fixed partition of LPR fails to handle such case neither.

\paragraph{Crowded obstacles} Row 4 shows an example of cluttered obstacles in traffic jams. Only a small portion of ground surface is visible to the lidar. MRF detects ground surface correctly but generates false-positives on vehicles (Box 4.1). Vanilla RANSAC misses one ground region (Box 4.2) as the fitted plane is polluted by false-positive ground points.

\paragraph{Low fences}
Box 5 shows another case where MRF fails. Similar to far obstacles, when there is no ground points visible in the neighborhood, MRF is confused whether the points are from low obstacles or sloped ground. Vanilla RANSAC also produces false-positives in Box 5 due to polluted inliers.

\paragraph{Non-horizontal ground}
We note that LPR fails to detect some obvious ground (Row 1, 2, 3 and 5). This is because the road is not perfectly horizontal. Lowest Point Representatives are all distributed in a small region at one side. Thus the plane overfitted from this small region may not fit the rest part of the ground.

\begin{table}
    \centering
    \vspace{1em}
    \begin{tabular}{c|c}
        \hline\hline
        & Runtime \\ \hline
        Proposed & $82$ms \\ 
        Vanilla RANSAC & $20$ms \\ 
        MRF \cite{Zhang2015} & $>1$s \\ 
        LPR \cite{Zermas2017} & $19$ms \\ \hline
    \end{tabular}
    \caption{Average runtime comparison}
    \label{tab:runtime}
\end{table}

\subsection{Runtime Comparison}
Table \ref{tab:runtime} compares the average runtime of the four approaches in our experiments. Vanilla RANSAC and LPR similarly have the best time efficiency. Our proposed approach introduces acceptable computational burden caused by tangent computation, integral image computation and partition search. MRF is the slowest with similar runtime as reported in \cite{Zhang2015}. Note that all approaches can be further accelerated via SIMD or GPU.

\section{Conclusions}
Our approach proposes two techniques to enhance the ground surface detection in sparse point cloud regions. Tangent based inlier verification naturally applies to RANSAC schemes for point cloud captured by lidars with different beam resolution. Disjoint multiple plane fitting adaptively partitions ground surface s.t. segments can be optimally fitted by disjoint planes. We show that the proposed approach effectively improves ground surface detection for sparse point cloud.

\bibliographystyle{IEEEtranSN}
\bibliography{main}

\end{document}